\documentclass[10pt,twocolumn,letterpaper]{article}

\usepackage[pagenumbers]{cvpr} 

\usepackage{multirow}
\usepackage{wrapfig}
\usepackage{appendix}
\usepackage[table,xcdraw]{xcolor}
\usepackage{booktabs}
\usepackage{amssymb}
\usepackage{amsmath}
\usepackage{mathtools}

\usepackage{tcolorbox}
\definecolor{lightgreen}{rgb}{0.8, 1, 0.8}
\definecolor{lightblue}{rgb}{0.8, 0.9, 1}
\definecolor{lightyellow}{rgb}{1, 1, 0.8}
\definecolor{blue1}{RGB}{220,235,250}
\definecolor{blue2}{RGB}{185,210,240}
\newcommand{\coolname}{$\texttt{VLAForge}$\xspace}

\definecolor{cvprblue}{rgb}{0.21,0.49,0.74}
\usepackage[pagebackref,breaklinks,colorlinks,allcolors=cvprblue]{hyperref}

\def\paperID{16510} 
\def\confName{CVPR}
\def\confYear{2025}

\title{Unleashing Vision-Language Semantics for Deepfake Video Detection}

\author{
  \textbf{Jiawen Zhu}\textsuperscript{1} \quad
  \textbf{Yunqi Miao}\textsuperscript{2} \quad
  \textbf{Xueyi Zhang}\textsuperscript{3} \quad
  \textbf{Jiankang Deng}\textsuperscript{4}\thanks{Corresponding authors: J. Deng (\texttt{j.deng16@imperial.ac.uk}) and G. Pang (\texttt{gspang@smu.edu.sg})} \quad
  \textbf{Guansong Pang}\textsuperscript{1}\footnotemark[1] \\
  \textsuperscript{1}Singapore Management University, Singapore\\
  \textsuperscript{2}The University of Warwick, UK\\
  \textsuperscript{3}Nanyang Technological University, Singapore\\
  \textsuperscript{4}Imperial College London, UK\\
}

\begin{document}
\vspace{-0.5cm}
\maketitle

\begin{abstract}
Recent Deepfake Video Detection (DFD) studies have demonstrated that pre-trained Vision-Language Models (VLMs) such as CLIP exhibit strong generalization capabilities in detecting artifacts across different identities.
However, existing approaches focus on leveraging visual features only, overlooking their most distinctive strength — the rich vision-language semantics embedded in the latent space.
We propose \coolname, a novel DFD framework that unleashes the potential of such cross-modal semantics to enhance model's discriminability in deepfake detection.
This work i) enhances the visual perception of VLM through a \textbf{ForgePerceiver}, which acts as an independent learner to capture diverse, subtle forgery cues both granularly and holistically, while preserving the pretrained Vision–Language Alignment (VLA) knowledge,
and ii) provides a complementary discriminative cue — \textbf{Identity-Aware VLA score}, derived by coupling cross-modal semantics with the forgery cues learned by ForgePerceiver. 
Notably, the VLA score is augmented by an identity prior-informed text prompting to capture authenticity cues tailored to each identity, thereby enabling more discriminative cross-modal semantics.
Comprehensive experiments on video DFD benchmarks, including classical face-swapping forgeries and recent full-face generation forgeries, demonstrate that our \coolname substantially outperforms state-of-the-art methods at both frame and video levels. Code is available at \renewcommand\UrlFont{\color{blue}}\url{https://github.com/mala-lab/VLAForge}.

\end{abstract}
\vspace{-0.2cm}

\section{Introduction}
The rapid progress of generative models has made realistic facial forgeries increasingly accessible, posing serious security and ethical concerns.
This trend has motivated the development of robust deepfake detection methods.
Traditional research primarily focused on capturing spatial artifacts~\cite{wang2020cnn, li2020face, dang2020detection, nguyen2024laa} or temporal inconsistencies within visual content~\cite{zheng2021exploring, wang2023altfreezing, choi2024exploiting, xu2023tall}.
Despite efforts have been made to strengthen model's robustness against unseen manipulation~\cite{cao2022end, sun2022dual, larue2023seeable, cheng2024can, lin2024fake, yan2024transcending, shiohara2022detecting}, the improvements in cross-dataset generalization are limited due to the scarcity and limited diversity of training data.

Recently, Vision–Language Models (VLMs), such as CLIP~\cite{radford2021learning}, have demonstrated remarkable generalization across diverse visual tasks, owing to their large-scale pretraining that aligns visual and textual modalities in a unified semantic space~\cite{zhou2022conditional, zhu2025fine, nguyen2024laa}. Inspired by this, recent studies have explored adapting CLIP for deepfake video detection.
Most existing approaches, however, focus on enhancing the visual encoder itself by adapter-based tuning~\cite{khan2024clipping, cui2025forensics}, bias mitigation~\cite{fu2025exploring}, or spatiotemporal modeling~\cite{han2025towards, yan2025generalizing}, while overlooking the intrinsic textual–visual aligned semantics within VLMs.
\begin{figure}[t]
    \centering
    \includegraphics[width=\linewidth]{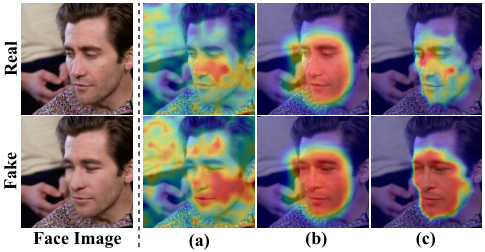}
    \caption{Visualization of \textbf{(a)} visual attention map of CLIP, \textbf{(b)} forgery localization map of ForgePerceiver, and \textbf{(c)} VLA attention map. Without proper adaptation, CLIP focuses on task-irrelevant visual cues. ForgePerceiver improves this case by highlighting potential forgery areas, but it provides coarse spatial guidance only. Augmented by discriminative identity priors, the VLA attention map offers more fine-grained, stronger forgery indication. 
    }
    \vspace{-0.1cm}
    \label{fig:intro}
    \vspace{-0.5cm}
\end{figure}
In this work, we propose \coolname to unleash the potential of cross-modal semantics to enhance the discriminability of VLMs in deepfake detection.

Generally, manipulated facial regions often exhibit diverse and heterogeneous artifacts, \eg, boundary inconsistencies, texture distortions. However, these low-level yet highly informative cues cannot be effectively captured by the VLM’s visual encoder, as these VLMs are primarily trained to understand the semantic objects in an image rather than to detect the artifacts in their pretraining.
As a result, when applied to deepfake detection, they often distribute their attention to different possible objects, which are typically not associated with forgeries, as illustrated in Fig.~\ref{fig:intro} (\textbf{a}).
To address the issue, we propose \textbf{ForgePerceiver}, which acts as an independent learner in \coolname to capture diverse and subtle forgery cues both granularly and holistically, while preserving the pretrained knowledge within VLM. This is achieved by simultaneously learning \textbf{i)} a set of diverse, subtle forgery-aware masks that modulate visual tokens from the VLM to model global authenticity,
and \textbf{ii) }a forgery localization map, which provides coarse region-aware 
cues that indicate the authenticity of the face region, as shown in Fig.~\ref{fig:intro} (\textbf{b})-\textbf{Bottom}.

Although \textbf{ForgePerceiver} works better than the primary VLMs in capturing the visual forgery evidence, this visual-only modality provides only coarse guidance on potential forgery regions, still lacking the ability to well differentiate the fake samples from the real ones, as illustrated by the two samples in Fig.~\ref{fig:intro} (\textbf{b}).
To mine fine-grained forgery cues, we further exploit the intrinsic visual–language alignment (VLA) within the VLMs, aiming to adapt the VLMs to learn discriminative patch-level authenticity cues from a cross-modal semantics perspective. While there have been some prior approaches \cite{lin2025standing,sun2025towards} that reprogram the VLMs to the DFD task, they focus on only high-level (\ie, image-level) visual–language alignment.
To tackle this challenge, we further introduce another \coolname component, \textbf{Identity-Aware VLA Scoring}. Its key insight is that, by injecting discriminative identity (ID) priors into the text prompts, the textual-visual alignment is adapted to be more fine-grained, enabling the model to capture authenticity cues tailored to each individual. The resulting VLA attention map can effectively highlight facial forgery regions with higher spatial precision when applied to fake samples, while refraining from exerting false attentions on real samples, as shown in Fig.~\ref{fig:intro} (\textbf{c}). 
To enable the VLA score to capture both coarse- and fine-grained authenticity cues, this VLA attention map is coupled with the forgery localization map from ForgePerceiver to enforce semantically grounded deepfake detection. Our contributions are summarized as:

\begin{itemize}
\item We propose a novel video DFD framework \coolname. Beyond merely refining visual representations as in existing VLM-based methods, \coolname unleashes the potential of cross-modal semantics to enhance the VLM's discriminability in deepfake detection.

\item \coolname consists of two novel components: ForgePerceiver and Identity-Aware VLA Scoring. The former learns to modulate the visual tokens from the VLM granularly and holistically, thereby capturing diverse subtle forgery cues; the latter provides a discriminative cue by leveraging cross-modal semantics via the ID prior-informed text prompts and the visual forgery cues from ForgePerceiver.

\item Comprehensive experiments on nine diverse DFD datasets, including both face-swapping and full-face generation forgeries, show that \coolname substantially outperforms state-of-the-art methods at both frame and video levels under cross-dataset settings.
\end{itemize}

\section{Related Work}
\noindent{\textbf{Deepfake Video Detection.}}
Traditional deepfake video detection approaches primarily rely on physiological inconsistencies such as unnatural eye blinking or mismatched head poses~\cite{li2018ictu, yang2019exposing, haliassos2021lips}, as well as identity-related inconsistencies that exploit semantic mismatches between facial regions and contextual cues~\cite{nirkin2021deepfake, dong2022protecting, dong2023implicit, bai2023aunet, huang2023implicit}. Another major line of research focuses on identifying universal forgery artifacts in the spatial and frequency domains~\cite{wang2020cnn, li2020face, dang2020detection, nguyen2024laa}, as well as on spatiotemporal modeling that captures temporal inconsistencies across frames~\cite{zheng2021exploring, wang2023altfreezing, choi2024exploiting, xu2023tall}. To further improve cross-dataset generalization, recent works incorporate contrastive or reconstruction learning~\cite{cao2022end, sun2022dual}, one-class detection formulations~\cite{larue2023seeable}, and synthetic or augmented data strategies~\cite{cheng2024can, lin2024fake, yan2024transcending, shiohara2022detecting}. This evolution from low-level visual cues to more semantic and generalizable representations naturally paves the way for exploring VLMs with high-level multimodal priors for authenticity reasoning.

\noindent{\textbf{Pre-trained VLMs in Deepfake Video Detection.}}
Recent advances in deepfake video detection have explored adapting pre-trained VLMs, such as CLIP~\cite{radford2021learning}, to leverage their strong generalization capability.
Visual adaptation methods, including CLIPping~\cite{khan2024clipping}, UDD~\cite{fu2025exploring}, ForAda~\cite{cui2025forensics}, FCG~\cite{han2025towards}, and Yan \etal ~\cite{yan2025generalizing}, mainly tune the visual encoder through adapters, temporal decoders, or bias-mitigation strategies to enhance feature discriminability and cross-dataset robustness.
However, these approaches are restricted to visual-only modeling and fail to exploit the semantic alignment inherently available in VLMs. To incorporate semantic priors, RepDFD~\cite{lin2025standing} reprograms pre-trained VLMs by generating sample-specific text prompts conditioned on external face embeddings, but the alignment between visual and textual features relies solely on global cosine similarity, lacking 
fine-grained (\eg, patch-level) authenticity correspondence and supervision.
FFTG~\cite{sun2025towards} enriches DFD datasets using synthesized image-text pairs with face region masks. The model's interpretability is enhanced by additional textual descriptions rather than adapting the visual–text alignment within VLMs.
In contrast, our \coolname enhances CLIP’s discriminability in deepfake detection by sharpening its intrinsic cross-modal semantics and diverse forgery cues from an independent authenticity learner.

\section{Preliminaries}
\noindent\textbf{Problem Statement.}
Deepfake Video Detection (DFD) aims to train a model that determines whether a given face video is authentic or fake. Formally, let the training dataset be $\mathcal{D}_{train} = \{\mathbf{v}_i, \mathbf{y}_i, \mathbf{G}_i\}_{i=1}^N$, where each video consists of $K$ frames, denoted as $\mathbf{v}_i = \{x_i^1, \dots, x_i^K\}$. Each frame $x_i^k \in \mathbb{R}^{3 \times h \times w}$ is an RGB image with spatial resolution $h \times w$. The video-level label $\mathbf{y}_i \in \{0, 1\}$ indicates whether $v_i$ is fake ($\mathbf{y}_i = 1$) or real ($\mathbf{y}_i = 0$), and consequently all frames within $\mathbf{v}_i$ inherit the same label $\mathbf{y}_i$. When available, $\mathbf{G}_i = \{G_i^1, \dots, G_i^K\}$ denotes the corresponding pixel-level forgery masks of frames in $\mathbf{v}_i$ for supervision. 

Since deepfake videos are created from diverse identities and generative models, DFD methods are commonly evaluated in a cross-dataset setting, where a model trained on $\mathcal{D}_{train}$ is tested on a set of target datasets $\mathcal{T} = \{\mathcal{D}_{test}^1, \mathcal{D}_{test}^2, \dots, \mathcal{D}_{test}^t\}$ containing identities and generation methods unseen during training. Given a query frame $x$ from a video $\mathbf{v}$, our objective is to learn a DFD model that produces an authenticity score $s(x)$, assigning higher values to fake frames and lower values to real ones.

\noindent\textbf{VLM Backbone.}
We build our framework upon CLIP~\cite{radford2021learning}, a vision–language model (VLM) demonstrating promising effectiveness in deepfake detection recently. CLIP consists of a text encoder $f_t(\cdot)$ and a visual encoder $f_v(\cdot)$, with the text and image representations from these encoders well aligned by pre-training on web-scale text-image pairs. Typically, $f_v(\cdot)$ comprises $L$ ViT block layers. The output of each block layer comprises a class token embedding $\mathbf{z}$ and patch token embeddings $\mathbf{P}$, which respectively encode the global and local visual information of the input image.

\section{Methodology}
\subsection{Overview}
\begin{figure*}[t]
    \centering
    \includegraphics[width=0.9\linewidth]{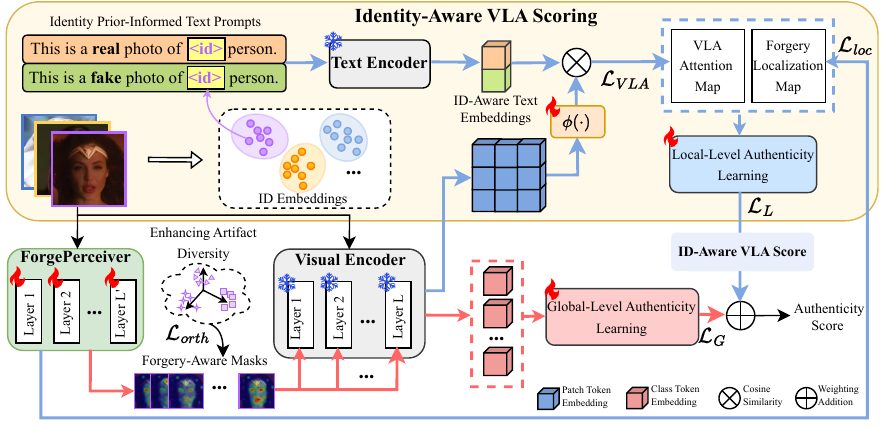}
    \caption{Overview of \coolname. It exploit the potential of VLMs in deepfake detection by i) ForgePerceiver, which acts as an independent learner to capture diverse, subtle forgery cues granularly and holistically; and ii) Identity-Aware VLA Scoring, which is driven by identity prior-informed text prompting and its enriched cross-modal semantics coupled with the visual forgery cues from ForgePerceiver.}
    \vspace{-0.2cm}
    \label{fig:overview}
    \vspace{-0.3cm}
\end{figure*}

In this work, we propose a novel framework \coolname for deepfake video detection. The goal of \coolname is to unleash the potential of cross-modal semantics embedded in VLMs to enhance the model’s discriminability in identifying facial forgeries. As illustrated in Fig.~\ref{fig:overview}, \coolname includes two key components, \textbf{ForgePerceiver} and \textbf{Identity-Aware VLA Scoring}, which jointly enable comprehensive and robust cross-modal authenticity reasoning. 
Specifically, ForgePerceiver operates independently of the VLM and serves as a specialized visual forgery learner that captures diverse, subtle artifact cues both granularly and holistically. It achieves this by simultaneously produces i) a set of diverse, subtle forgery-aware masks that modulate visual tokens of VLM's visual encoder $f_v(\cdot)$, enabling a holistic learning of global-level authenticity; and ii) a forgery localization map that provides coarse region-aware forgery cues that indicates the authenticity of the face region.

On the other hand, the Identity-aware VLA Scoring module aims to adapt the VLM to learn discriminative patch-level authenticity cues from
a cross-modal semantics perspective. It first enriches the text prompts with identity priors to produce more discriminative ID-aware text embeddings, which are then compared with the patch token embeddings from $f_v(\cdot)$ to derive a VLA attention map through vision–language alignment. The resulting map is coupled with the forgery localization map from ForgePerceiver to enforce semantically grounded deepfake detection.
Below we present these components in detail.

\subsection{ForgePerceiver}
To enhance the visual perception capability of VLMs on task-specific data without compromising its inherent pretrained knowledge, we introduce ForgePerceiver, which independently learns to capture diverse, subtle forgery cues and adapts this information to effectively empower the VLMs. ForgePerceiver adopts a lightweight ViT architecture that operates on two types of tokens: visual tokens, obtained by image patch embeddings from the pretrained VLM; and learnable query tokens introduced to probe different forgery priors.
Through the interactions between the query and visual tokens, ForgePerceiver generates two complementary forms of forgery priors: a set of diverse, subtle forgery-aware masks for holistic, global-level reasoning, and a forgery localization map capturing coarse region-aware spatial cues to support local-level reasoning.

\noindent\textbf{Learning of Forgery-Aware Masks.}
To exploit the pretrained recognition capability of the VLM, we adopt the class token in $f_v(\cdot)$ as the global representation of each query sample. However, this global embedding is typically insensitive to subtle forgery artifacts.
To mitigate this limitation, ForgePerceiver derives diverse, subtle forgery-aware masks based on different query tokens and use them to modulate the VLM's class token, enriching its global representation with artifact-specific cues. 

Formally, let $\mathbf{V} \in \mathbb{R}^{h_v \times w_v \times d_v}$ and $\mathbf{Q} \in \mathbb{R}^{q \times d_v}$ respectively denote the visual tokens and the learnable query tokens, where $h_v \times w_v$ represents the spatial patch grid, $d_v$ is the embedding dimension, and $q$ is the number of learnable query tokens $\mathbf{Q}$. To enable effective interaction, both tokens are projected into task-specific feature spaces through learnable mappings, yielding $\hat{\mathbf{V}} = g_1(\mathbf{V}) \in \mathbb{R}^{h_v \times w_v \times (d_v \times H)}$, $\hat{\mathbf{Q}} = g_2(\mathbf{Q}) \in \mathbb{R}^{q \times d_v}$, where $H$ denotes the number of attention heads in $f_v(\cdot)$.
For each $i$-th attention head, a forgery-aware mask is computed by measuring the similarity between each query token and the corresponding head-specific visual features:
\begin{equation}
\begin{aligned}
\mathcal{M}_{i} = \hat{\mathbf{Q}} \, {\hat{\mathbf{V}}_i^{\top}},
\qquad i=1, \dots, H.
\end{aligned}
\label{eq:localization}
\end{equation}

This results in a set of $H$ head-specific forgery-aware masks, denoted as $\mathcal{M} = \{\mathcal{M}_i\}_{i=1}^H \in \mathbb{R}^{H \times q \times h_v \times w_v}$.
To ensure that different queries capture complementary artifact priors rather than the redundant cues, we first average the head-specific masks across the head dimension to obtain query-wise forgery-aware masks $\hat{\mathcal{M}} = \frac{1}{H} \sum_{i=1}^{H} \mathcal{M}_{i} \in \mathbb{R}^{q \times h_v \times w_v}$.
We then impose an orthogonality constraint on $\hat{\mathcal{M}}$ to explicitly encourage diversity among the artifact priors learned by different queries:
\begin{equation}
\mathcal{L}_{orth} = \sum_{u \neq v}^{q} \left|
\frac{\Psi(\hat{\mathcal{M}}_{u}) \cdot \Psi(\hat{\mathcal{M}}_{v})}
{|\Psi(\hat{\mathcal{M}}_{u})| , |\Psi(\hat{\mathcal{M}}_{v})|}
\right|,
\label{eq:dis}
\end{equation}
where $\Psi(\cdot)$ denotes the vectorization operation, and $\hat{\mathcal{M}}_{j} \in \mathbb{R}^{h_v \times w_v}$ represents the forgery-aware mask corresponding to the $j$-th query.

\noindent\textbf{Global-Level Authenticity Learning.}
The forgery-aware masks $\mathcal{M}$ are subsequently integrated into the self-attention mechanism of $f_v(\cdot)$, guiding the attention distribution so that the class tokens accumulate more discriminative, forgery-aware semantics. To enable this, the class token is replicated into $q$ instances, each paired with a corresponding query-wise mask to attend differently to patch tokens according to their associated artifact priors.

Formally, let $\mathbf{Z}_{cls} = \{\mathbf{z}_1, \dots \mathbf{z}_q\}$ denote the replicated class token embeddings. In the $l$-th ViT block of the VLM, the query matrix for the $j$-th class token is denoted by $\mathbb{Q}_j^{(l)}$, and the key and value matrices derived from the patch tokens $\mathbf{P}$ are denoted by $\mathbb{K}_{P}^{(l)}$ and $\mathbb{V}_{P}^{(l)}$, respectively. Then the update of the $j$-th class token  in the $i$-th attention head is computed by injecting the attention bias $\hat{\mathcal{M}}_{i,j}$:
\begin{equation}
\mathbf{z}_{j}^{(l)} = \mathrm{softmax}\left(\frac{\mathbb{Q}_j^{(l)}{\mathbb{K}_{P}^\top}^{(l)}}{\sqrt{d}} + \mathcal{M}_{i,j}\right)\mathbb{V}_{P}^{(l)}.
\end{equation}

After being refined through $L$-layer forgery-aware attention, the enriched class-token representations are fed into an authenticity scoring head $\eta_1(\cdot)$ to produce a global-level authenticity score $s_g = \eta_1(\mathbf{Z}_{cls}^{(L)})$, which is optimized by a binary classification loss:
\begin{equation}
\mathcal{L}_{G} = \frac{1}{N}\sum_{x \in X_{train}} 
\mathcal{L}_{ce}\big(s_g, y_x\big),
\end{equation}
where $\mathcal{L}_{ce}$ is the cross-entropy loss and $y_x \in \{0,1\}$ represents the ground-truth label of $x$.

\noindent\textbf{Forgery Localization.}
ForgePerceiver further incorporates a Forgery Localization task that supplies auxiliary spatial guidance, enabling the model to learn more accurate artifact priors without sacrificing their diversity. The localization map delivers coarse region-aware artifact evidence that are further used for local-level authenticity learning.

Specifically, we employ another projection function $g_3(\cdot)$ to transform the visual tokens $\mathbf{V}$ into a task-adaptive feature space for forgery localization, \ie, $\tilde{\mathbf{V}} = g_3(\mathbf{V}) \in \mathbb{R}^{h_v \times w_v \times d_v}$.
The query-wise forgery localization map $\{\tilde{\mathcal{M}}\}_{j=1}^q \in \mathbb{R}^{q \times h_v \times w_v}$ is computed following Eq.~\ref{eq:localization}, where the projected feature $\tilde{\mathbf{V}}$ is used in place of $\hat{\mathbf{V}}$. To obtain the final localization map, the query-wise maps are aggregated using a convolutional head $h(\cdot)$: \begin{equation}
\begin{aligned}
\mathbf{M}_{loc} = h([\tilde{\mathcal{M}}_{1}, \dots, \tilde{\mathcal{M}}_{q}]),
\end{aligned}
\label{eq:merge_localization}
\end{equation}
where $[\cdot]$ denotes channel-wise concatenation along the query dimension.
The objective for the forgery localization can then be defined as:
\begin{equation}
\begin{aligned}
\mathcal{L}_{loc} = \frac{1}{N}\sum_{x \in X_{train}} \mathcal{L}_{mse}\left({\Phi_{loc}(\mathbf{M}}_{loc}), G_x\right),
\end{aligned}
\label{eq:loss_local}
\end{equation}
where $\Phi_{loc}(\cdot)$ is an interpolation function that unsamples the forgery localization map $\mathbf{M}_{loc} \in \mathbb{R}^{h_v \times w_v}$ to the image resolution $(h, w)$, $\mathbf{G}_x$ is the ground-truth forgery mask of frame $x$, and $\mathcal{L}_{mse}(\cdot)$ denotes the mean squared error loss.

\subsection{Identity-Aware VLA Scoring}
Relying solely on low-level visual cues restricts prior VLM-based detectors from leveraging the intrinsic vision–language alignment available within VLMs. Such cross-modal semantics provide complementary perspectives and can serve as valuable patch-level authenticity indicators. To this end, \coolname introduces an Identity-aware VLA Scoring module to model fine-grained forgery cues by exploiting such semantics. It first enriches text prompts with discriminative identity priors and generates a VLA attention map by aligning them with patch-token embeddings from $f_v(\cdot)$. This resulting map is then fused with the forgery localization map from ForgePerceiver, yielding a discriminative ID-aware VLA score that effectively complements the global-level authenticity reasoning.

\noindent\textbf{Identity Prior-Informed Text Prompting.}
To facilitate effective vision–language alignment, it is crucial to encode authenticity semantics within the text modality. Following the standard prompt design adopted in CLIP-based detection methods~\cite{lin2025standing, zhu2024generalistanomalydetectionincontext, zhou2023anomalyclip, zhu2025fine}, \coolname first constructs a text prompt pair describing real and fake faces. To further obtain identity-aware textual representations, \coolname introduces an identity prior-informed text prompting strategy that explicitly incorporates an identity prior-based token into these templates. Specifically, the text prompts in \coolname is formulated as:
\begin{equation}
\begin{aligned}
    \mathcal{T}^r &= \text{``This is a real photo of \textless id\textgreater\ person.''}; \\
    \mathcal{T}^f &= \text{``This is a fake photo of \textless id\textgreater\ person.''},
\end{aligned}
\label{eq:prompt}
\end{equation}
where $\textless id\textgreater$ denotes a placeholder token, and $r$ and $f$ indicate the real and fake classes, respectively. After tokenization, each prompt $\mathcal{T}^c$ ($c \in \{r, f\}$) is transformed into token embeddings $\mathbf{T}^c =[T^c_1, T^c_2, \dots, T^c_{|\mathbf{T}^c|}]^\top \in \mathbb{R}^{|\mathbf{T}^c| \times d_{tk}}$, where $|\mathbf{T}^c|$ is the sequence length of the prompt tokens and $d_{tk}$ is the dimensionality of each token embedding.

Let $\tau$ denote the index of the $\textless id\textgreater$ in the tokenized prompt.
For each query frame $x$, \coolname refines its textual token embeddings by substituting the embedding at position $\tau$ with the corresponding class token embedding $\mathbf{z}^{(L)}$ obtained from the final ViT block of $f_v(\cdot)$:
\begin{equation}
\begin{aligned}
\hat{\mathbf{T}}^c =
\begin{cases}
\hat{T}^c_i = \mathbf{z}^L, & \text{if } i = \tau^c,\\[2pt]
\hat{T}^c_i = T^c_i, & \text{otherwise},
\end{cases}
\qquad \ c\in\{r,f\}.
\end{aligned}
\label{eq:refine_prompt}
\end{equation}

The refined textual token embeddings are then fed into the VLM's text encoder $f_t(\cdot)$ to obtain the ID-aware text features, $\mathbf{F}_r \in \mathbb{R}^{d_t}$ and $\mathbf{F}_f \in \mathbb{R}^{d_t}$, corresponding to the real and fake classes, respectively.

\noindent\textbf{Learning of VLA Attention Map.}
To leverage the intrinsic visual–language alignment of VLMs for the DFD task, \coolname generates an VLA attention map by measuring the similarity between the ID-aware text features $\{\mathbf{F}_r,\mathbf{F}_f\}$ and patch-token embeddings of a given frame produced by $f_v(\cdot)$. Specifically, let $\mathbf{P} \in \mathbb{R}^{h_p \times w_p \times d_p}$ denote the patch-token embeddings, where $h_p$ and $w_p$ represent the spatial dimensions of the patch grid, and $d_p$ is the embedding dimension,
the attention score of the VLA attention map $\mathbf{M}_{VLA} \in \mathbb{R}^{h_p \times w_p}$ at spatial location $(i,j)$ can then be computed by:
\begin{equation}
\begin{aligned}
\mathbf{M}_{VLA}(i,j) =
\frac{\exp\big(\phi(\mathbf{P}(i,j))\mathbf{F}_f^{\top}\big)}
{\sum\limits_{c \in \{r,f\}} \exp\big(\phi(\mathbf{P}(i,j))\mathbf{F}_{c}^{\top}\big)},
\end{aligned}
\label{eq:similarity-map}
\end{equation}
where $\mathbf{P}(i,j)$ is the the corresponding patch-token embedding at location $(i,j)$, $(\cdot)^{\top}$ denotes the transpose operation, and $\phi(\cdot)$ is a learnable adapter that projects each patch embedding from the visual feature dimension $d_p$ to the text feature dimension $d_t$. To adapt the visual–language alignment more effectively to the deepfake detection task, we supervise $\mathbf{M}_{VLA}$ against the corresponding ground-truth forgery mask $\mathbf{G}_x$, which is formulated as:
\begin{equation}
\begin{aligned}
\mathcal{L}_{VLA} = \frac{1}{N}\sum_{x \in X_{train}} \mathcal{L}_{Dice}\left({\Phi_{VLA}(\mathbf{M}}_{VLA}), \mathbf{G}_x\right),
\end{aligned}
\label{eq:loss_id}
\end{equation}
where $\Phi_{VLA}(\cdot)$ is an interpolation operator that upsamples the attention map $\mathbf{M}_{VLA}$ to the image resolution, and $\mathcal{L}_{Dice}(\cdot)$ denotes the dice loss. 

\noindent\textbf{Local-Level Authenticity Learning.}
In addition to the discriminative capability provided by holistic class-token representations, the local-level discriminative information from coarse region-aware evidence, \ie, forgery localization map $\mathbf{M}_{loc}$ and VLA attention map $\mathbf{M}_{id}$, are also significant and serve as complementary knowledge to the global-level authenticity learning. To combine these complementary sources of evidence, we apply element-wise fusion and subsequently synthesize these maps through a learnable fusion network $\psi(\cdot)$, which can be formulated as:
\begin{equation}
\begin{aligned}
\mathbf{F}_{x} = \psi(\mathbf{M}_{loc} \odot \mathbf{M}_{id}),
\end{aligned}
\label{eq:feature_fusion}
\end{equation}
where $\odot$ denotes the element-wise multiplication. The fused feature $\mathbf{F}_{x}$ is subsequently fed into an authenticity scoring head $\eta_2(\cdot)$, which produces the ID-aware VLA score $s_{VLA} = \eta_2(\mathbf{F}_{x})$. The score is then optimized by a binary classification loss:
\begin{equation}
\begin{aligned}
    \mathcal{L}_{L} = \frac{1}{N}\sum_{x\in X_{train}} \mathcal{L}_{ce}(s_{VLA}, y_x).
\end{aligned}
\label{eq:loss_b1}
\end{equation}
\begin{table*}[t]
\centering
\caption{AUROC results of  frame-level and video-level deepfake detection. $^{\dag}$ indicates results reproduced by us. The best and second-best results are respectively highlighed in \protect\colorbox{lightblue}{\textbf{blue}} and \colorbox{lightyellow}{yellow}.}
\vspace{-0.1cm}
\resizebox{0.9\textwidth}{!}{
\begin{tabular}{cccccc|ccccc}
\toprule
\multicolumn{6}{c|}{\textbf{Frame-level AUROC}}                   & \multicolumn{5}{c}{\textbf{Video-level AUROC}}                                                                                                                                          \\
\textbf{Method}                             & \textbf{Venue}                  & \textbf{CDF-v1}                      & \textbf{CDF-v2}                      & \textbf{DFDC}                        & \textbf{DFD}                         & \textbf{Method}                       & \textbf{Venue}               & \textbf{CDF-v2}                      & \textbf{DFDC}                       & \textbf{DFD}                         \\ \midrule
\multicolumn{1}{c|}{\textbf{Xception}~\cite{rossler2019faceforensics++}}      & \multicolumn{1}{c|}{ICCV’19}    & 77.9                                 & 73.7                                 & 70.8                                 & 81.6                                 & \multicolumn{1}{c|}{\textbf{TALL}~\cite{xu2023tall}}    & \multicolumn{1}{c|}{ICCV’23} & 83.1               & 69.3                     & 83.3                                     \\
\multicolumn{1}{c|}{\textbf{EfficientB4}~\cite{luo2023beyond}}   & \multicolumn{1}{c|}{ICML‘19}    & 79.1                                 & 74.9                                 & 69.6                                 & 81.5                                 & \multicolumn{1}{c|}{\textbf{SeeABLE}~\cite{larue2023seeable}} & \multicolumn{1}{c|}{ICCV’23} & 87.3                                 & 75.9                                 & -                                    \\
\multicolumn{1}{c|}{\textbf{X-ray}~\cite{li2020face}}         & \multicolumn{1}{c|}{CVPR'20}    & 70.9                                 & 67.9                                 & 63.3                                 & 76.6                                 & \multicolumn{1}{c|}{\textbf{IID}~\cite{huang2023implicit}}     & \multicolumn{1}{c|}{CVPR’23} & 83.8                                 & 70.0                                   & 93.9                                   \\
\multicolumn{1}{c|}{\textbf{FFD}~\cite{dang2020detection}}           & \multicolumn{1}{c|}{CVPR'20}    & 78.4                                 & 74.4                                 & 70.3                                 & 80.2                                 & \multicolumn{1}{c|}{\textbf{SFDG}~\cite{wang2023dynamic}}    & \multicolumn{1}{c|}{CVPR’23} & 75.8                                 & 73.6                                 & 88.0                                 \\
\multicolumn{1}{c|}{\textbf{SPSL}~\cite{liu2021spatial}}          & \multicolumn{1}{c|}{CVPR'21}    & 81.5                                 & 76.5                                 & 70.4                                 & 81.2                                 & \multicolumn{1}{c|}{\textbf{CADDM}~\cite{dong2023implicit}}   & \multicolumn{1}{c|}{CVPR’23} & 93.9                                 & 73.9                                 & -                                    \\
\multicolumn{1}{c|}{\textbf{SRM}~\cite{luo2021generalizing}}           & \multicolumn{1}{c|}{CVPR'21}    & 79.3                                 & 75.5                                 & 70.0                                 & 81.2                                 & \multicolumn{1}{c|}{\textbf{LAA-NET}~\cite{nguyen2024laa}} & \multicolumn{1}{c|}{CVPR’24} & 95.4                                 & -                                    & 80.0                                 \\
\multicolumn{1}{c|}{\textbf{Recce}~\cite{cao2022end}}         & \multicolumn{1}{c|}{CVPR’22}    & 76.8                                 & 73.2                                 & 71.3                                 & 81.2                                 & \multicolumn{1}{c|}{\textbf{SAM}~\cite{choi2024exploiting}}     & \multicolumn{1}{c|}{CVPR’24} & 89.0                                 & -                                    & 96.1                                 \\
\multicolumn{1}{c|}{\textbf{UCF}~\cite{yan2023ucf}}           & \multicolumn{1}{c|}{ICCV’23}    & 77.9                                 & 75.3                                 & 71.9                                 & 80.7                                 & \multicolumn{1}{c|}{\textbf{Yan \textit{et al.}}~\cite{yan2025generalizing}}  & \multicolumn{1}{c|}{CVPR’25} & 94.7                                 & 84.3                                 & \cellcolor{lightyellow}{96.5}                                    \\
\multicolumn{1}{c|}{\textbf{ED}~\cite{yan2024effort}}        & \multicolumn{1}{c|}{AAAI'24}     & 81.8                                   & 86.4                               & 72.1                                & -                                    & \multicolumn{1}{c|}
{\textbf{FCG}~\cite{han2025towards}}  & \multicolumn{1}{c|}{CVPR’25} & 95.0                                 & 81.8                                 & -                                    \\
\multicolumn{1}{c|}{\textbf{UDD}~\cite{fu2025exploring}}           & \multicolumn{1}{c|}{AAAI’25}    & -                                    & 86.9                                 & 75.8                                 & 91.0                                 & \multicolumn{1}{c|}
{\textbf{Effort}~\cite{yan2024effort}}  & \multicolumn{1}{c|}{ICML'25}  & 95.6                                 & 84.3                                 & 96.5                                   \\
\multicolumn{1}{c|}{\textbf{SBI}~\cite{shiohara2022detecting}}           & \multicolumn{1}{c|}{CVPR’22}    & 83.1                                    & 80.2                                 & 71.4                                 & 77.4                                 & \multicolumn{1}{c|}{\textbf{SBI}~\cite{shiohara2022detecting}}     & \multicolumn{1}{c|}{CVPR’22} & 93.2                                 & 72.4                                 & 88.2                                 \\
\multicolumn{1}{c|}{\textbf{ProDet}~\cite{cheng2024can}}          & \multicolumn{1}{c|}{Neurips’24}    & 90.9                                 & 84.5                                 & 72.4                                 & -                                 & \multicolumn{1}{c|}{\textbf{ProDet}~\cite{cheng2024can}}    & \multicolumn{1}{c|}{Neurips’24} & 92.6                                 & 70.7                                 & 90.1                                 \\
\multicolumn{1}{c|}{\textbf{LSDA}~\cite{yan2024transcending}}          & \multicolumn{1}{c|}{CVPR’24}    & 86.7                                 & 83.0                                 & 73.6                                 & 88.0                                 & \multicolumn{1}{c|}{\textbf{LSDA}~\cite{yan2024transcending}}    & \multicolumn{1}{c|}{CVPR’24} & 89.8                                 & 73.5                                 & 95.6                                 \\
\multicolumn{1}{c|}{\textbf{CDFA}~\cite{lin2024fake}}        & \multicolumn{1}{c|}{ECCV'24}     & -                                    & 89.9                                 & 78.7                                 & -                                    & \multicolumn{1}{c|}{\textbf{CDFA}~\cite{lin2024fake}}  & \multicolumn{1}{c|}{ECCV'24}  & 93.8                                 & 83.0                                 & 95.4                                    \\
\multicolumn{1}{c|}{\textbf{RepDFD}~\cite{lin2025standing}}        & \multicolumn{1}{c|}{AAAI'25}     & 83.0                                    & 80.0                                 & 77.3                                 & 85.8$^{\dag}$                                    & \multicolumn{1}{c|}{\textbf{RepDFD}~\cite{lin2025standing}}  & \multicolumn{1}{c|}{AAAI'25}  & 89.9                                 & 81.0                                 & 95.1$^{\dag}$                                    \\
\multicolumn{1}{c|}{\textbf{ForAda}~\cite{cui2025forensics}}        & \multicolumn{1}{c|}{CVPR’25}    & \cellcolor{lightyellow}{91.4}                                 & \cellcolor{lightyellow}{90.0}                                 & \cellcolor{lightyellow}{84.3}                                 & \cellcolor{lightyellow}{93.3}                                 & \multicolumn{1}{c|}{\textbf{ForAda}~\cite{cui2025forensics}}  & \multicolumn{1}{c|}{CVPR’25} & \cellcolor{lightyellow}{95.7}                                 & \cellcolor{lightyellow}{87.2}                                 & \cellcolor{lightyellow}{96.5}$^{\dag}$ \\
\multicolumn{1}{c|}{\textbf{\coolname}}          & \multicolumn{1}{c|}{-}          &  \cellcolor{lightblue}{\textbf{93.9}} & \cellcolor{lightblue}{\textbf{91.2}} & \cellcolor{lightblue}{\textbf{87.0}} & \cellcolor{lightblue}{\textbf{93.6}} & \multicolumn{1}{c|}{\textbf{\coolname}}    & \multicolumn{1}{c|}{-}       & \cellcolor{lightblue}{\textbf{96.8}} & \cellcolor{lightblue}{\textbf{89.6}} & \cellcolor{lightblue}{\textbf{97.2}} \\ \bottomrule
\end{tabular}
}
\label{main_results}
\vspace{-0.3cm}
\end{table*}

Therefore, the overall learning objective of \coolname is as follows:
\begin{equation}
\mathcal{L}_{final} = \mathcal{L}_{loc} + \mathcal{L}_{VLA} + \mathcal{L}_{G} + \mathcal{L}_{L}.
\end{equation}

During inference, the final authenticity score for a query frame $x'$ is obtained by combining the predictions from the global- and local-level authenticity reasoning branches:
\begin{equation}
s(x') = \alpha s_g' + (1-\alpha)s_{VLA}',
\label{final_score}
\end{equation}
where $\alpha$ is a hyperparameter that balances the contributions of the global-level score $s_g'$ and the local-level score $s_{VLA}'$.

\section{Experiments}
\noindent\textbf{Datasets.} 
We evaluate our method on five classical face-swapping forgery datasets: FaceForensics++ (FF++)~\cite{rossler2019faceforensics++}, CelebDF v1/v2~\cite{li2020celeb} (CDF-v1/v2), Deepfake Detection Challenge (DFDC)~\cite{dolhansky2020deepfake}, and DeepfakeDetection (DFD)~\cite{deepfake2021}; as well as full-face synthesized data based on CDF-v2 sourced from the large-scale DF40 dataset~\cite{yan2024df40}, where we select five representative GAN and Diffusion-based generative models: VQGAN~\cite{esser2021taming}, StyleGAN-XL (StyleGAN)~\cite{sauer2022stylegan}, SiT-XL/2 (SiT)~\cite{atito2021sit}, DiT~\cite{peebles2023scalable}, and PixArt~\cite{chen2023pixart} (see \texttt{Appendix} A for details about the datasets).

\noindent\textbf{Evaluation Protocol and Metrics.}
To assess the generalization ability, we follow the common cross-dataset evaluation protocol by training the model on the c23-compression version of FF++ and evaluating it on the remaining datasets.
Following previous deepfake video detection studies~\cite{cheng2024can, yan2024transcending, lin2024fake, cui2025forensics, lin2025standing}, we adopt the Area Under the Receiver Operating Characteristic (AUROC) as the primary evaluation metric for both frame-level and video-level performance.
For video-level evaluation, each query video is decomposed into a sequence of frames, and the final video-level score is then obtained by averaging the frame-level predictions.

\noindent\textbf{Implementation Details.}
The implementation details and complexity analysis for \coolname and competing methods are provided in \texttt{Appendix} B. and \texttt{Appendix} C.1.

\subsection{Comparison with State-of-the-Art Methods}
\noindent\textbf{Generalization to Classical Forgery Faces.}
Table~\ref{main_results} (\textbf{Left}) presents the frame-level cross-dataset results of \coolname compared with 16 state-of-the-art (SotA) methods across four face forgery benchmarks. Overall, \coolname outperforms all competing approaches across all datasets. To be specific, the VLM-based approaches such as ForAda and UDD achieve better cross-dataset performance compare to non-VLM-based methods such as LSDA and CDFA, benefiting from the superior VLM generalization capability. By more effectively exploiting the multimodal recognition capacity of the VLMs through the proposed ForgePerceiver and Identity-aware VLA Scoring, \coolname achieves substantial performance gains, particularly on the largest and most diverse DFDC dataset. As a result, \coolname surpasses the second-best methods by up to 2.7\% AUROC.

Additionally, Table~\ref{main_results} (\textbf{Right}) shows the video-level cross-dataset comparison against 16 SotA methods across three datasets. We exclude CDF-v1 from comparison since it is a smaller subset of CDF-v2 and has been rarely reported in recent works. In general, \coolname consistently surpasses all competing methods across datasets, achieving up to 2.4\% AUROC improvement over the best competing method. These consistent gains further validate the effectiveness of \coolname in enhancing the discriminative power and generalization capability of the VLM for deepfake video detection.
\begin{table}[t]
\centering
\caption{AUROC results on frame (F)- and video (V)-level detection of GAN- and diffusion-generated full-face forgeries. 
}
\vspace{-0.1cm}
\resizebox{0.9\linewidth}{!}{
\begin{tabular}{c|c|ccccc}
\toprule
\textbf{Setting} & \textbf{Method} & \textbf{VQGAN} & \textbf{StyleGAN} & \textbf{SiT} & \textbf{DiT} & \textbf{PixArt} \\ 
\midrule
\multirow{3}{*}{\textbf{F-level}} 
    & \textbf{RepDFD} 
        & 80.5 & 82.2 & 58.2 & 58.7 & 88.5 \\
    & \textbf{ForAda} 
        & 93.9 
        & 92.5
        & 69.0
        & 62.0
        & 96.5 \\
    & \textbf{Ours} 
        & \cellcolor{lightblue}{\textbf{98.4}}
        & \cellcolor{lightblue}{\textbf{98.0}}
        & \cellcolor{lightblue}{\textbf{77.4}}
        & \cellcolor{lightblue}{\textbf{70.7}}
        & \cellcolor{lightblue}{\textbf{97.2}} \\
\midrule
\multirow{3}{*}{\textbf{V-level}} 
    & \textbf{RepDFD} 
        & 85.0 & 86.3 & 69.3 & 58.9 & 94.8 \\
    & \textbf{ForAda} 
        & 98.1
        & 97.9
        & 76.0
        & 67.1
        & 98.5 \\
    & \textbf{Ours} 
        & \cellcolor{lightblue}{\textbf{99.7}}
        & \cellcolor{lightblue}{\textbf{99.7}}
        & \cellcolor{lightblue}{\textbf{85.9}}
        & \cellcolor{lightblue}{\textbf{80.3}}
        & \cellcolor{lightblue}{\textbf{99.5}} \\
\bottomrule
\end{tabular}
}
\label{GAN_Diffusion_results}
\vspace{-0.2cm}
\end{table}

\noindent\textbf{Generalization to Full-Face Generated Deepfakes.}
To further assess the generalization of \coolname, we extend our evaluation on more challenging fully-face generation datasets, where the forgery faces are synthesized by advanced GAN- and diffusion-based generators. Unlike face-swapping forgeries that contain visible blending artifacts, these data exhibit coherent appearance and high-fidelity details.
As shown in Table~\ref{GAN_Diffusion_results}, we compare \coolname with two SotA methods, RepDFD\footnote{As no official implementation of RepDFD is available, we reproduce it for the comparison and will release our implementation publicly.}~\cite{lin2025standing} and ForAda~\cite{cui2025forensics}.
The results show that \coolname significantly outperforms all baselines across the datasets, demonstrating its effectiveness in capturing intrinsic generative traces and identity-related cues, thereby achieving strong transferability to diverse and highly realistic synthesis scenarios. 

\begin{table}[t]
\centering
\caption{Frame (F)- and video (V)-level module ablation results.}
\vspace{-0.1cm}
\resizebox{0.9\linewidth}{!}{
\begin{tabular}{c|c|cccccc}
\toprule
\textbf{Setting} & \textbf{Model} & \textbf{CDF-v2} & \textbf{DFDC}  & \textbf{DFD}   & \textbf{VQGAN} & \textbf{SiT} \\ 
\midrule
\multirow{5}{*}{\textbf{F-level}} 
    & \textbf{Base}   & 58.3 & 64.0 & 77.5 & 74.8 & 52.9 \\
    & \textbf{+ T1}   & 76.3 & 76.0 & 74.6 & 89.7 & 69.3 \\
    & \textbf{+ T2}   & 82.3 & 80.9 & 87.4 & 95.1 & 74.6 \\
    & \textbf{+ T3}   & 90.8 & 86.5 & 92.8 & 97.6 & 76.8 \\
    & \textbf{+ T4}   & \cellcolor{lightblue}{\textbf{91.2}} & \cellcolor{lightblue}{\textbf{87.0}} & \cellcolor{lightblue}{\textbf{93.6}} & \cellcolor{lightblue}{\textbf{98.4}} & \cellcolor{lightblue}{\textbf{77.4}} \\
\midrule
\multirow{5}{*}{\textbf{V-level}} 
    & \textbf{Base}   & 60.3 & 64.6 & 83.9 & 79.4 & 59.2 \\
    & \textbf{+ T1}   & 78.9 & 80.6 & 76.5 & 98.9 & 79.9 \\
    & \textbf{+ T2}   & 87.9 & 83.1 & 93.1 & 98.6 & 83.1 \\
    & \textbf{+ T3}   & 96.1 & 89.4 & 96.7 & 99.4 & 84.9 \\
    & \textbf{+ T4}   & \cellcolor{lightblue}{\textbf{96.8}} & \cellcolor{lightblue}{\textbf{89.6}} & \cellcolor{lightblue}{\textbf{97.2}} & \cellcolor{lightblue}{\textbf{99.7}} & \cellcolor{lightblue}{\textbf{85.9}} \\
\bottomrule
\end{tabular}
}
\label{ablation_study}
\vspace{-0.5cm}
\end{table}

\subsection{Analysis of \coolname}
\noindent\textbf{Module Ablation.}
We conduct an ablation study on \coolname’s two core modules, ForgePerceiver and the ID-aware VLA Scoring module, by progressively enabling their constituent configurations on top of the baseline. The results across five datasets are summarized in Table~\ref{ablation_study}, where the baseline (`\textbf{Base}') denotes a frozen CLIP model that performs deepfake detection by computing the similarity between the class token embedding and text features derived from simple handcrafted prompts. 

To validate the contribution of ForgePerceiver, `\textbf{+T1}' introduces local-level authenticity learning based on the forgery localization map generated by this module. Building upon this, `\textbf{+T2}' further enhances the global-level authenticity reasoning by incorporating the forgery-aware masks from ForgePerceiver into the self-attention mechanism of CLIP’s visual encoder.

For the Identity-aware VLA Scoring module, `\textbf{+T3}' incorporates the VLA attention map to strengthen local-level authenticity reasoning, while `\textbf{+T4}' further refines the textual representations by integrating identity priors into the prompts. The resulting ID-aware VLA attention map provides more discriminative supervision and achieves consistent improvements across all datasets, forming the complete \coolname framework. The consistent improvement from `\textbf{Base}' to `\textbf{+T4}' demonstrates that each components in \coolname provides complementary gains and contributes to stronger robustness and cross-dataset generalization. More results and analysis about module ablation can be found in \texttt{Appendix} C.2.

\begin{table}[t]
\centering
\caption{
Frame (F)- and video (V)-level loss ablation results.
}
\vspace{-0.1cm}
\resizebox{0.9\linewidth}{!}{
\begin{tabular}{c|cc|ccccc}
\toprule
\textbf{Setting} & $\mathcal{L}_{VLA}$ & $\mathcal{L}_{orth}$ 
& \textbf{CDF-v2} & \textbf{DFDC} & \textbf{DFD} & \textbf{VQGAN} & \textbf{SiT} \\
\midrule

\multirow{4}{*}{\textbf{F-level}}
& $\times$ & $\times$ 
    & 88.4 & 84.0 & 91.0 & 96.2 & 74.8 \\
& $\times$ & \checkmark 
    & 89.1 & 84.7 & 92.1 & 97.2 & 75.0 \\
& \checkmark & $\times$ 
    & \cellcolor{lightyellow}{91.0} 
    & \cellcolor{lightyellow}{86.9}
    & \cellcolor{lightyellow}{92.8}
    & \cellcolor{lightyellow}{97.3}
    & \cellcolor{lightyellow}{76.4} \\
& \checkmark & \checkmark 
    & \cellcolor{lightblue}{\textbf{91.2}}
    & \cellcolor{lightblue}{\textbf{87.0}}
    & \cellcolor{lightblue}{\textbf{93.6}}
    & \cellcolor{lightblue}{\textbf{98.4}}
    & \cellcolor{lightblue}{\textbf{77.4}} \\
\midrule

\multirow{4}{*}{\textbf{V-level}}
& $\times$ & $\times$
    & 93.6 & 86.6 & 95.1 & 97.8 & 85.0 \\
& $\times$ & \checkmark
    & 93.9 & 87.3 & 96.1 
    & \cellcolor{lightyellow}{99.2}
    & \cellcolor{lightyellow}{85.1} \\
& \checkmark & $\times$
    & \cellcolor{lightyellow}{96.5}
    & \cellcolor{lightyellow}{89.3}
    & \cellcolor{lightyellow}{96.5}
    & 97.9 & 84.4 \\
& \checkmark & \checkmark
    & \cellcolor{lightblue}{\textbf{96.8}}
    & \cellcolor{lightblue}{\textbf{89.6}}
    & \cellcolor{lightblue}{\textbf{97.2}}
    & \cellcolor{lightblue}{\textbf{99.7}}
    & \cellcolor{lightblue}{\textbf{85.9}} \\
\bottomrule
\end{tabular}
}
\label{loss_ab}
\vspace{-0.2cm}
\end{table}

\begin{figure}[h]
    \centering
    \includegraphics[width=0.9\linewidth]{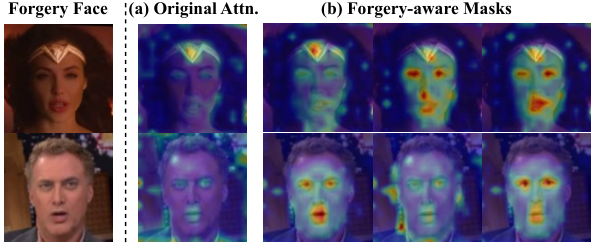}
    \caption{Attention visualization of forgery faces produced by different models: \textbf{(a)} Attention from original CLIP; \textbf{(b)} Attention of forgery-aware masks from ForgePrecever.
    }
    \vspace{-0.3cm}
    \label{fig:query_map}
\end{figure}

\noindent\textbf{Loss Ablation.}
We further analyze the impact of the two loss terms in \coolname: $\mathcal{L}_{VLA}$, which guides the learning of VLA attention maps, and $\mathcal{L}_{orth}$, which encourages diversity among the forgery-aware masks.
As shown in Table~\ref{loss_ab}, by jointly optimizing these two objectives, \coolname effectively learns complementary identity-aware and artifact-diverse priors, achieving the best performance across all datasets. Removing $\mathcal{L}_{VLA}$ leads to a noticeable performance drop, as the model can no longer anchor ID-aware textual semantics to manipulated regions in the VLA attention maps. This disrupts the adaptation of vision–language alignment semantics to the deepfake detection task. 
\begin{figure*}[h]
    \centering
    \includegraphics[width=0.95\linewidth]{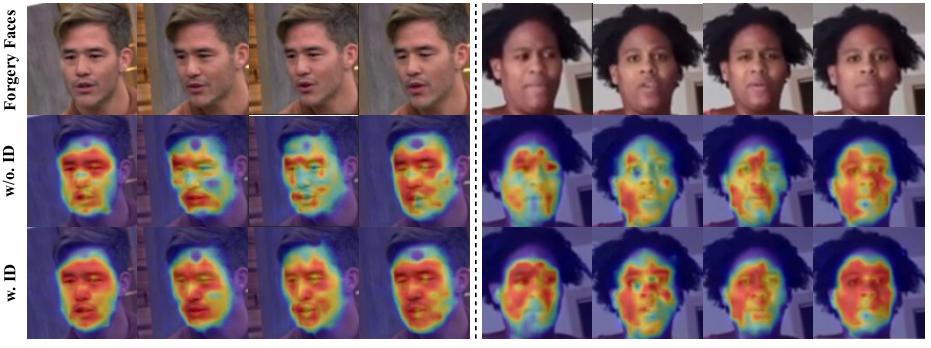}
    \vspace{-0.3cm}
    \caption{Visualization of VLA attention maps with (w.) and without (w/o.) injecting identity prior into text prompts.}
    \label{fig:id_map}
    \vspace{-0.2cm}
\end{figure*}
On the other hand, removing $\mathcal{L}_{orth}$ yields a more significant decline on fully synthesized deepfakes. This is because, without this constraint, the forgery-aware masks generated by ForgePrecever collapse into redundant artifact patterns, impairing the model’s discriminability to capture heterogeneous generative traces. Fig.~\ref{fig:query_map} provides a more concrete illustration of the effectiveness of $\mathcal{L}_{orth}$. With $\mathcal{L}_{orth}$, the forgery-aware masks learned by ForgePerceiver distribute their attention across distinct semantically meaningful facial regions (\eg, eyes, mouth, boundary), enabling the model to capture complementary forgery cues, while the attention map from original CLIP exhibit simple, non-discriminative forgery areas. 

\noindent\textbf{Effectiveness of Identity Priors.}
Fig~\ref{fig:id_map} visualizes the VLA attention maps produced by \coolname with and without injecting identity priors into the text prompts.
Without the identity prior, the model lacks identity-conditioned guidance and thus interprets each frame independently based on simple generic tokens (\eg, `real' or `fake') for the face semantics. As a result, the attention becomes sparse and inconsistent across frames, especially under variations in pose or facial expression. In contrast, with the ID-conditioned prompts, the attention becomes more spatially coherent across frames and more accurately highlights the complete forged facial region, enabling the cross-modal semantic reasoning to attend to more consistent, fine-grained authenticity cues. Additional attention and t-SNE visualization and analysis can be found in \texttt{Appendix} C.3.

\begin{figure}[t]
    \centering
    \includegraphics[width=0.95\linewidth]{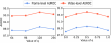}
    \caption{Frame-level and Video-level AUROC based on different value of $q$ (\textbf{Left}) and $\alpha$ (\textbf{Right}).
    }
    \vspace{-0.1cm}
    \label{fig:para}
    \vspace{-0.5cm}
\end{figure}
\noindent\textbf{Hyperparameter Sensitivity Analysis.}
We further investigate the sensitivity of \coolname to two key hyperparameters: the number of query tokens $q$ and the fusion weight $\alpha$ in Eq.~\ref{final_score}. The averaged results across five datasets are shown in Fig.~\ref{fig:para}. In particular, the model achieves the best performance when $q = 128$, while further increasing $q$ leads to a slight performance decline. This suggests that using too few queries restricts the model’s capacity to capture diverse forgery patterns, whereas an excessively large $q$ introduces noisy or biased variations rather than meaningful artifact cues, particularly under the $\mathcal{L}_{orth}$ constraint, which enforces strong diversity across query-wise forgery-aware masks.
Regarding the fusion weight $\alpha$, which balances the contributions of global-level and local-level authenticity scoring, the best performance is achieved near $\alpha = 0.5$.
This indicates that global semantic information and localized forgery cues play equally important and complementary roles in producing reliable authenticity predictions. More analyses can be found in \texttt{Appendix} C.4.

\vspace{-0.1cm}
\section{Conclusion}
In this work, we presented \coolname, a novel framework that enhances deepfake video detection by leveraging cross-modal semantics within VLM. \coolname introduces an \textbf{ForgePerceiver} to capture diverse, subtle  forgery cues without disrupting the pretrained VLA knowledge, and an \textbf{Identity-aware VLA Scoring} module that enriches textual prompts with identity priors to derive discriminative patch-wise authenticity guidance. By jointly modeling forgery-aware and identity-aware semantics, \coolname achieves strong complementary reasoning and substantially improves cross-dataset generalization. Extensive experiments on nine deepfake benchmarks, including face swapping and full-face generation, demonstrate its superiority over existing SotA methods.

\section*{Acknowledgment}
This research is partially supported by A*STAR under its MTC YIRG Grant (M24N8c0103), the Singapore Ministry of Education (MOE) Academic Research Fund (AcRF) Tier 1 Grant (24-SIS-SMU-008), and the Lee Kong Chian Fellowship (T050273). J. Deng was supported by the NVIDIA Academic Grant.

{
    \small
    \bibliographystyle{ieeenat_fullname}
    \bibliography{main}
}

\newpage
\appendix
\section{Dataset Details}
\subsection{Data Statistics of Training and Testing}
We evaluate our method on five classical face-swapping forgery datasets: FaceForensics++ (FF++)~\cite{rossler2019faceforensics++}, CelebDF v1/v2~\cite{li2020celeb} (CDF-v1/v2), Deepfake Detection Challenge (DFDC)~\cite{dolhansky2020deepfake}, and DeepfakeDetection (DFD)~\cite{deepfake2021}; as well as full-face synthesized data based on CDF-v2 sourced from the large-scale DF40 dataset~\cite{yan2024df40}, where we select five representative GAN and Diffusion-based generative models: VQGAN~\cite{esser2021taming}, StyleGAN-XL (StyleGAN)~\cite{sauer2022stylegan}, SiT-XL/2 (SiT)~\cite{atito2021sit}, DiT~\cite{peebles2023scalable}, and PixArt~\cite{chen2023pixart}.

To assess the generalization ability, we follow the common cross-dataset evaluation protocol by training the model on the c23-compression version of FF++ and evaluating it on the remaining datasets. Table~\ref{face-swap} provides the data statistics of classical face-swapping forgery datasets, while Table~\ref{full-face} shows the full-face synthesized datasets generated by GAN and Diffusion-based generative models.

\begin{table}[t!]
\centering
\caption{Data statistics of face-swapping forgery datasets.}
\resizebox{\linewidth}{!}{
\begin{tabular}{c|cccc}
\toprule
\textbf{Dataset}         & \textbf{Sythesis Methods} & \textbf{Real} & \textbf{Fake} & \textbf{Total} \\ \midrule
\textbf{FaceForensics++} & 4                         & 1000                 & 4000                 & 5000                  \\
\textbf{CelebDF v1}      & 1                         & 408                  & 795                  & 1203                  \\
\textbf{CelebDF v2}      & 1                         & 590                  & 5634                 & 6229                  \\
\textbf{DFDC}            & 8                         & 23654                & 104500               & 128154                \\
\textbf{DFD}             & 5                         & 363                  & 3000                 & 3363                  \\ \bottomrule
\end{tabular}}
\label{face-swap}
\end{table}

\begin{table}[t!]
\centering
\caption{Data statistics of full-face synthesized data generated based on GAN and Diffusion-based methods.}
\resizebox{\linewidth}{!}{
\begin{tabular}{c|cccc}
\toprule
\textbf{Dataset}     & \textbf{Synthesis Type} & \textbf{Real} & \textbf{Fake} & \textbf{Total} \\ \midrule
\textbf{VQGAN}       & GAN based               & 590                  & 5634                 & 6229                  \\
\textbf{StyleGAN-XL} & GAN based               & 590                  & 5634                 & 6229                  \\
\textbf{SiT-XL/2}    & Latent Diffusion        & 590                  & 5634                 & 6229                  \\
\textbf{DiT}         & Latent Diffusion        & 590                  & 5634                 & 6229                  \\
\textbf{PixArt}      & Latent Diffusion        & 590                  & 5634                 & 6229                  \\ \bottomrule
\end{tabular}}
\label{full-face}
\end{table}

\subsection{Classical Face-Swapping Forgery Datasets}
\noindent\textbf{FaceForensics++ (FF++)}~\cite{rossler2019faceforensics++}. 
FF++ is a widely used benchmark for facial manipulation detection, containing over 1,000 real videos and their corresponding manipulated versions generated with four representative face-swapping and reenactment techniques: DeepFakes, Face2Face, FaceSwap, and NeuralTextures. Multiple compression levels are also provided to simulate real-world media quality.

\noindent\textbf{CelebDF v1/v2 (CDF-v1/v2)}~\cite{li2020celeb}.
Celeb-DF is a large-scale deepfake video dataset constructed using YouTube celebrity videos and high-quality swapping methods designed to reduce visual artifacts. Version v2 significantly improves visual realism compared to v1, making it more challenging for detection models.

\noindent\textbf{Deepfake Detection Challenge (DFDC)}~\cite{dolhansky2020deepfake}.
DFD contains high-quality deepfake videos created with professional actors under controlled conditions. Compared with FF++ and DFDC, DFD offers cleaner visual quality and fewer compression artifacts, providing an ideal benchmark for evaluating fine-grained detection capability.
 
\noindent\textbf{DeepfakeDetection (DFD)}~\cite{deepfake2021}.
This dataset is designed for deepfake detection tasks, providing a comprehensive collection of video sequences that can be used to train and evaluate deep learning models for identifying manipulated media. It was downloaded from the official FaceForensics server, which offers high-quality datasets specifically for the purpose of face manipulation detection.

\subsection{Full-Face Synthesized Datasets}
With the progress of AIGC techniques, full-face synthesis has achieved high perceptual realism without typical blending artifacts found in face-swapping methods. We evaluate on five representative GAN- and diffusion-based generators from DF40~\cite{yan2024df40}, covering different generative families and image priors. DF40 includes fully generated subsets derived from both CelebDF-v2 and FF++. Because our model is trained on FF++, we adopt the subset generated from CelebDF-v2 for cross-dataset evaluation to ensure non-overlapping identities and generation patterns.

\noindent\textbf{VQGAN}~\cite{esser2021taming}.
A GAN-based discrete latent-space generator capable of producing high-resolution images with improved perceptual quality. It synthesizes globally coherent facial structures without explicit patch-level inconsistencies.

\noindent\textbf{StyleGAN-XL (StyleGAN)}~\cite{sauer2022stylegan}.
An improved variant of StyleGAN capable of scaling to diverse large-scale datasets with strong identity realism, making generated faces more diverse and visually convincing.

\noindent\textbf{SiT-XL/2 (SiT)}~\cite{atito2021sit}.
A diffusion-based generator that leverages scalable transformer architectures for high-fidelity face synthesis, producing smooth textures and natural facial layouts.

\noindent\textbf{DiT}~\cite{peebles2023scalable}.
A transformer-based diffusion model that operates directly in latent space. It provides high-quality generative realism with minimal local artifacts, further increasing detection difficulty.

\noindent\textbf{PixArt}~\cite{chen2023pixart}.
A recent high-resolution text-to-image generator demonstrating strong semantic alignment and photo-realism. The produced faces lack typical low-level cues, posing challenges to artifact-based detectors.

\section{Implementation Details}
\subsection{Details of Model Configuration}
We implement \coolname using OpenCLIP with the publicly available \textit{ViT-L/14} backbone. The parameters of both the visual and text encoders in CLIP are kept frozen throughout all experiments.
The ForgePerceiver follows the \textit{vit\_tiny\_patch16\_224} configuration, with all parameters randomly initialized and fully trained from scratch during optimization. 
The numbers of forgery query tokens and replicated class-token embeddings $q$ are set to 128 by default. The number of fusion weight $\alpha$ is set to 0.5. We adopt the Adam optimizer with an initial learning rate of 2e-5 and a weight decay of 5e-4 to update model parameters. The input images are resized to 224$\times$224, and the batch size is set to 32. To ensure that the model learns to recognize both real and fake faces across diverse identities while mitigating overfitting, training is conducted for 15 epochs on a single NVIDIA GeForce RTX 3090 GPU. We will release the code upon publication to facilitate reproducibility.

\subsection{Implementation of Comparison Methods}
\noindent\textbf{ForAda}~\cite{cui2025forensics}.
ForAda enhances CLIP for face forgery detection by introducing a task-specific adapter that learns forgery-related visual traces and interacts with CLIP’s visual tokens while preserving its inherent generalization capability. The results on classical face-swapping datasets are taken from the original paper, whereas the results on full-face generation datasets are reproduced using the official implementation\footnote{https://github.com/OUC-VAS/ForensicsAdapter}.

\noindent\textbf{RepDFD}~\cite{lin2025standing}.
Since RepDFD does not provide official code, we reproduce the method based on the implementation details described in the paper. We adopt CLIP-ViT-L/14 pretrained on LAION-400M as the foundation model, with an input resolution of 224 $\times$ 224 and an input transformation parameter of $p = 34$. We employed the AdamW optimizer with
the learning rate 1.0, and the weight decay was fixed at 0. Besides, the data preprocessing transform was as same as the original CLIP, and the visual
prompt was initialized by zero. For the external identity-embedding network, we follow the paper and employ a pre-trained TransFace model~\cite{dan2023transface}. We also experiment with ArcFace~\cite{papantoniou2024arc2face}, but both models occasionally produce invalid or empty embeddings due to low-quality or non-face input regions. We will release our implementation publicly.

\section{Additional Results}
\subsection{Model Complexity of \coolname vs. SotA Methods}
Table~\ref{complexity} presents a comparison of model complexity and video-level AUROC across several representative deepfake detection approaches. The results highlight a clear performance–efficiency advantage of \coolname. To be specific, traditional CNN- and transformer-based detectors (\eg, DCL~\cite{sun2022dual}, FTCN~\cite{zheng2021exploring}, CFM~\cite{luo2023beyond}) contain 19–26M parameters, yet their average AUROC remains limited. This reflects their weak generalization when applied to unseen datasets such as CDF-v2 and DFDC. RepDFD~\cite{lin2025standing} shows the smallest parameter count, but its performance indicates that extremely lightweight models generally sacrifice discriminability, especially in challenging real-world settings. ForAda~\cite{cui2025forensics} achieves a stronger balance between model size and performance (5.7M parameters, 91.5 AVG) due to its effective adapter-based design.

In contrast, \coolname uses only 3.28M parameters—smaller than ForAda and significantly smaller than most baselines—yet achieves the highest AUROC on both CDF-v2 (96.8) and DFDC (89.6), yielding the best overall average (93.2). This demonstrates that \coolname provides superior generalization and discriminative power with notably lower parameter complexity, validating the effectiveness of coupling cross-modal semantics with compact forgery-aware learning.

We also report training and inference time comparisons (mean±std) in Table~\ref{complexity}. As shown, our method requires slightly longer runtime than ForAda, but is more efficient than RepDFD. Notably, with only marginal overhead, \coolname achieves notably improvement in performance.

\begin{table}[t!]
\centering
\caption{Model complexity analysis in Video-level AUROC.}
\resizebox{\linewidth}{!}{
\begin{tabular}{c|ccc|ccc}
\hline
\textbf{Methods} & \textbf{Param.} & \textbf{Training (ms)} & \textbf{Inference (ms)} & \textbf{CDF-v2} & \textbf{DFDC} & \textbf{AVG}  \\ \hline
\textbf{DCL}     & 19.35M          & \textbf{-} & \textbf{-} & 82.3            & 76.7          & 79.5          \\
\textbf{FTCN}    & 26.6M          & \textbf{-}  & \textbf{-} & 86.9            & 74.0          & 80.5          \\
\textbf{CFM}     & 25.37M        & \textbf{-}   & \textbf{-} & 89.7            & 80.2          & 85.0          \\
\textbf{RepDFD}  & 0.078M        & \textbf{1832.4±7.5}  & \textbf{42.3±0.6}  & 89.9            & 81.0          & 85.5          \\
\textbf{ForAda}  & 5.7M           & \textbf{933.3±1.4} & \textbf{21.2±0.2}  & 95.7            & 87.2          & 91.5          \\ \hline
\textbf{Ours}    & 3.28 M         & \textbf{1410.0±2.2} & \textbf{41.7±0.9}  & \textbf{96.8}   & \textbf{89.6} & \textbf{93.2} \\ \hline
\end{tabular}}
\label{complexity}
\end{table}

\subsection{Text Prompting Variants Comparison}
To verify the importance of ID prior-informed text prompts in \coolname, we evaluate several prompt variants: i) replacing the backbone of CLIP with LLAVA (`LLAVA'); ii)replacing the simple prompts with LLM-generated descriptive prompts (`LLM-Prompts'); iii) replacing CLIP-based identity priors with those extracted using Arc2Face (`Arc2Face-ID'); iv) substituting fixed prompts with learnable prompts (Learnable-Prompt); and v) removing the generic tokens (i.e., `real/fake') from the prompts (w/o `Real/Fake'). As shown in Table~\ref{textprompt}, all variants lead to noticeable performance degradation.

The variant of `LLAVA' performs slightly worse than CLIP-based results. We attribute this to different training objectives: LLaVA is primarily optimized for visual instruction tuning tasks, while VLAForge benefits more from visual–language alignment, which is more di-015
rectly supported by CLIP’s contrastive pretraining. The performance drop with `LLM-Prompts' suggests that such prompts may lack consistent applicability across samples and exhibit weaker alignment with CLIP's visual representations of facial artifacts. In contrast, the simpler prompts in \coolname provide more stable visual-language alignment. The degradation of `Arc2Face-ID' likely stems from the fact that face recognition models focus on identity-discriminative features, which are less compatible with CLIP's text embedding space. Moreover, they are more sensitive to image quality, reducing robustness under low-quality or synthetic conditions. The decline in `Learnable-Prompt' indicates that, although learnable tokens introduce flexibility, they compromise semantic stability, leading to less robust alignment across identities and artifact types. In contrast, fixed prompts serve as consistent semantic anchors that better support identity-aware modulation. Finally, removing the generic tokens (`w/o Real/Fake') results in a significant performance drop, demonstrating that these tokens act as explicit textual anchors that enhance discriminative capability, rather than introducing circular reasoning.

\begin{table}[t!]
\centering
\caption{Frame (F)- and video (V)-level AUROC using different prompt variants.}
\resizebox{\linewidth}{!}{
\begin{tabular}{c|c|ccccc}
\toprule
\textbf{}                          & \textbf{Method}                      & \textbf{CDF-v2} & \textbf{DFDC}  & \textbf{DFD}   & \textbf{VQGAN} & \textbf{SiT} \\ \midrule
\multirow[c]{5}{*}{\textbf{F-level}} & \textbf{LLaVA}                & 89.8  & 85.8 & 92.1 & 98.1 & 76.6                         \\
                                 & \textbf{LLM-Prompt} & 89.5  & 86.3 & 92.7 & 97.5 & 75.1                         \\ 
& \textbf{Arc2Face-ID}                & 90.6  & 86.4 & 92.5 & 98.0 & 76.9                         \\
                                 & \textbf{Learn-Prompt} & 83.8  & 82.3 & 90.1 & 96.6 & 76.4                         \\
                                  &  \textbf{$w/o$ Real/Fake} & 89.5  & 86.3 & 92.7 & 97.5 & 75.1                         \\ \cmidrule{2-7}
                                 & \textbf{\coolname}                        & \textbf{91.2}  & \textbf{87.0} & \textbf{93.6}  & \textbf{98.4} & \textbf{77.4}                        \\ \midrule
\multirow[c]{6}{*}{\textbf{V-level}} & \textbf{LLaVA}                & 95.3  & 88.4 & 96.2 & 99.3 & 84.2                         \\
                                 & \textbf{LLM-Prompt} & 94.9  & 89.0 & 96.7 & 99.0 & 84.2                         \\
& \textbf{Arc2Face-ID}                & 95.9  & 89.0 & 96.8 & 99.4 & 84.4                         \\
                                 & \textbf{Learn-Prompt} & 89.7  & 85.1 & 94.4 & 98.7 & 84.0                         \\
                                 & \textbf{$w/o$ Real/Fake} & 94.9  & 89.0 & 96.7 & 99.0 & 84.2                         \\ \cmidrule{2-7}
                                 & \textbf{\coolname}                        & \textbf{96.8}  & \textbf{89.6} & \textbf{97.2} & \textbf{99.7} & \textbf{85.9}                         \\ \bottomrule
\end{tabular}}
\label{textprompt}
\end{table}

\subsection{Visualization Comparison}
\begin{figure}[t]
    \centering
    \includegraphics[width=\linewidth]{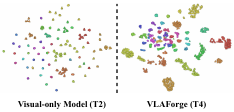}
    \caption{T-sNE visualization comparison of the DFD features between visual-only model (`\textbf{T2}' in Table 3) and complete \coolname (`\textbf{T4}' in Table 3).
    }
    \label{fig:tsne}
\end{figure}

\noindent\textbf{t-SNE Visualization.} A clear distinction can be observed between the feature distributions of `\textbf{T2}' and `\textbf{T4}'. As shown in Fig.~\ref{fig:tsne}, without leveraging cross-modal semantics, the visual-only features learned by `\textbf{T2}' fail to form meaningful clusters—samples do not exhibit identity-driven grouping, and real (positive) and fake (negative) samples lack a clear decision boundary, reflecting weak discriminability. In contrast, `\textbf{T4}' produces identity-consistent cluster structures, where real samples form compact clusters while fake samples are relatively scattered, indicating richer heterogeneity in forgery artifacts. This demonstrates that \coolname facilitates more discriminative, separable, and semantically well-organized feature representations.

\begin{figure}[t]
    \centering
    \includegraphics[width=\linewidth]{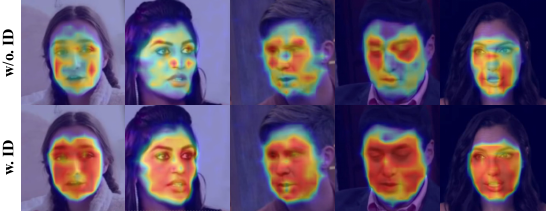}
    \caption{More visualization of VLA attention maps with (w.) and without (w/o.) injecting identity prior into text prompts.).
    }
    \label{fig:app}
\end{figure}

\noindent\textbf{VLA attention maps.}
Fig~\ref{fig:app} provides additional qualitative results illustrating VLA attention maps across a broader set of identities. Unlike the examples in the main text—where multiple frames from the same identity were compared to show temporal consistency—each identity here is represented by a single frame. Nonetheless, a consistent pattern emerges: without identity priors (top row), the attention maps remain sparse and unstable, often highlighting fragmented or irrelevant regions. In contrast, with identity priors injected into the text prompts (bottom row), the maps become substantially more coherent and identity-consistent, focusing on semantically meaningful facial regions. These results further validate that identity-conditioned textual semantics significantly enhance the spatial precision and reliability of VLA-based forgery indication.

\subsection{Hyperparameter Sensitivity Analysis}
Table~\ref{seed} reports the frame-level and video-level AUROC scores of \coolname under three different and commonly used random seeds to assess its robustness and training stability. Overall, the results demonstrate high consistency across seeds, with only minor performance fluctuations, confirming that the proposed method is not sensitive to random initialization. By default, we apply a fixed random seed of 1024 to ensure training stability and reproducibility.
\begin{table}[t!]
\centering
\caption{Frame (F)- and video (V)-level AUROC using different random seeds.}
\resizebox{\linewidth}{!}{
\begin{tabular}{c|cccccc}
\toprule
\multicolumn{1}{l|}{}             & \textbf{SEED} & \textbf{CDF-v2} & \textbf{DFDC} & \textbf{DFD}  & \textbf{VQGAN} & \textbf{SiT-XL/2} \\ \midrule
\multirow{3}{*}{\textbf{F-level}} & \textbf{1024}    & 91.2            & 87.0          & 93.6          & 98.3           & 77.4              \\
                                  & \textbf{0000}    & 91.4            & 87.9          & 92.8          & 98.6           & 76.9              \\
                                  & \textbf{1111}    & 90.7            & 86.7          & 93.2          & 97.9           & 77.6              \\ \cmidrule{2-7} 
\multicolumn{1}{l|}{}             & \textbf{AVG}  & \textbf{91.1}   & \textbf{87.2} & \textbf{93.2} & \textbf{98.3}  & \textbf{77.3}     \\ \midrule
\multirow{3}{*}{\textbf{V-level}} & \textbf{1024}    & 96.8            & 89.6          & 97.2          & 99.7           & 85.9              \\
                                  & \textbf{0000}    & 97.0            & 89.9          & 96.4          & 99.6           & 85.4              \\
                                  & \textbf{1111}    & 95.9            & 89.2          & 97.5          & 99.4           & 86.3              \\ \cmidrule{2-7} 
\multicolumn{1}{l|}{}             & \textbf{AVG}  & \textbf{96.5}   & \textbf{89.6} & \textbf{97.0} & \textbf{99.6}  & \textbf{85.8}     \\ \bottomrule
\end{tabular}}
\label{seed}
\end{table}

\end{document}